# DIVING PERFORMANCE ASSESSMENT BY MEANS OF VIDEO PROCESSING

Stefano Frassinelli[1], Alessandro Niccolai[1] and Riccardo E. Zich[1]

[1]Dipartimento di Energia, Politecnico di Milano, Milan, Italy
alessandro.niccolai@polimi.it

## ABSTRACT

*The aim of this paper is to present a procedure for video analysis applied in an innovative way to diving performance assessment. Sport performance analysis is a trend that is growing exponentially for all level athletes. The technique here shown is based on two important requirements: flexibility and low cost. These two requirements lead to many problems in the video processing that have been faced and solved in this paper.*

## 1. INTRODUCTION

Many trends are involving in these years sports [1]: the number of people involved in fitness activities is increasing for every level and for every age [2], [3].

In this euphoria for sports, performance analysis is becoming every year much more important. Athletes of all levels are involved in analysis of their performances. This is giving to many sport a push toward higher and higher level of the competitions [4],[5].

Diving is one of the sports involved in this trend. The performances of diving are quite difficult to be objectively and engineeringly measured, but this is really important both for the athlete that can have an understanding of the efficiency of their training, both for competition organizers that can have a tool to make the judgment more objective.

Some wearable devices have been introduced in literature [6], but they can be only used during training because they are not allowed during competitions.

In other sports, like soccer [7], video analysis have been introduced with generic techniques and with ad-hoc procedures [8].

Some studies have been done also for diving [9], but the techniques proposed are quite expansive for many low-level athlete applications.

In this paper a low-cost procedure for performance assessment in diving by means of video processing will be explained with some proper example of the results of this procedure.

The paper is structured as follows: in Section 2 the performance metrics in diving will be briefly introduced. In Section 3 the needs and the requirements that have driven the development of this technique are shown. Section 4 shows the proposed procedure and eventually Section 5 contains the conclusions.

## 2. PERFORMANCE METRICS

Performance estimation can change a lot from one sport to another [10]. In some sports the important quantities are obvious, while in others they are hard to be identified clearly.

Diving is a sport in which the evaluation is based in a strong way on the subjective judgment of experts. In any case, it is possible to identify some measurable quantities that are related with the final performance of the dive.

Identifying the metrics is important because it gives a preferential path to the procedure to follow for the analysis of the dive.

The metrics here identified and studied are:

•	Barycentre trajectory during the dive: this is an important metric [11] to understand possible mistakes made by the athlete during the jump, like lateral movements induced by a non-symmetric jump on the springboard [12], [13] or on the platform;

•	Barycentre position during the entrance in the water: this is a metric that is important because it is one of the evaluation parameters that are commonly used in competitions;

•	Maximum barycentre height [14]: this parameter is important because it gives an information of the time at disposal of the athlete for making all the figures that are required for the jump [15], [16].

These are not all the performance metrics that are used during competitions, but for sake of simplicity in this paper the analysis will be performed only of the barycentre positions. Other analysis can be done for examples on the forces transmitted to the platform or to the springboard.

In the next sections the needs and the requirements that have driven the development of this technique will be explained.

## 3. NEED AND REQUIREMENTS

Needs and Requirements analysis is a fundamental step in technique designing [17]. This is useful to frame the rage of applications and also to give a reasonable direction to the design phase. Moreover, it is important to finally assess the performance of the designed product, in this case the performance analysis technique.

The first, most important need is the flexibility of the technique: it should be used for almost any kind of video, for 10m platform diving and for 3m springboard diving.

These two possibilities are quite different because, while for 3m spring-board it is possible to imagine to have a fixed camera, for the 10m it is much more difficult: to have a good video it is necessary to place the camera far from the swimming pool and this leads to two problems: the first one is the space available. Again, for flexibility issues, the technique has to be suitable for the use in different places, so it must not require too much space. The second problem related is the perspective problem: the diminution of the size of the diver can be a problem when it must be isolated from the background.

Another consequence of the flexibility requirement, is that the vide can be done also without a specific equipment, so it can have vibrations.

The second need, is the requirement of a technique that can be applied also for non-expert diver, so it must be cheap. It is not possible to imagine to have a high cost equipment.

Next section provides a general description of the overall procedure, then the main steps will be deeply analysed.

## 4. VIDEO PROCESSING PROCEDURE

Video processing is the technique applied here to performance assessment. While also other techniques can be applied [6], them are limited by regulatory issues in competitions and by comfort of the athlete.

Video analysis is an effective technique [9] because it does not require any additional item on the divers that can influence performance (the psychological approach is really important especially during high level competitions [12]) or that are forbidden [18].

It is important that the proposed technique can be applied both in training and during competitions: in this way, it is possible to compare the response of the athlete to the stress.

Another important advantage of video processing is that it can be applied to official video of competitions (Olympic Games or other International Competitions). These videos can be used as benchmark in the training.

Moreover, video processing results can be combined with kinematic simulations [19] to give to trainers and to athlete a complete tool in training.

In the next subsection, the overall process flow chart is explained; in the following ones, the steps are described providing some example of their application.

**4.1. Process flow chart**

The overall process, described in Figure 1, is composed by five steps and it is aimed to pass from a video, acquired during training or found on the Internet, to a performance score. In this paper the first four steps are described.

The first step is the image extraction by sampling correctly the video. To have a correct sampling, some parameters have to be chosen properly. Section 4.2 shows the concepts that are behind this choice.

The second and the third steps are the core of the procedure of image processing: firstly, a panorama is created by mosaicking, then the barycentre of the diver is found in each image by properly apply a colour filter. These two steps are described respectively in Section 4.3 and 4.4.

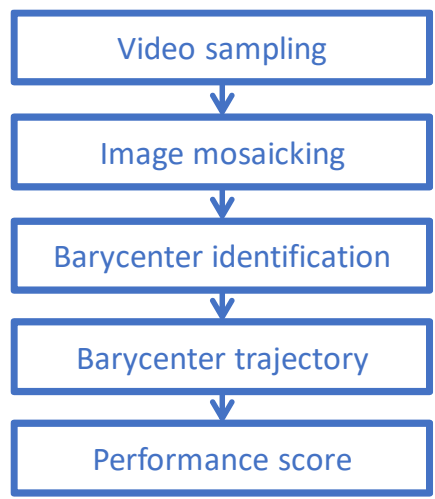

Figure 1.  Process flow chart

Having found the barycentre of the diver it is possible to finally reconstruct his trajectory and finally to associate a score to the dive.

**4.2. Video Sampling**

The first step of the procedure is the extraction of the frames from the video. This procedure is easy because all the videos are defined as a sequence of frames. Only a free variable remains that is the sampling frequency.

This factor is defined by the compromise between calculation time and results precision.

Regarding the calculation time, it is possible to make some considerations: the diving time is defined by the height of the platform. In the case of a 10m platform, it is possible to firstly analyse a dive without any initial speed to have an idea of the total time:

$$\Delta t^2 = \frac{\Delta x}{2g} = 0.5 \ [s] \tag{1}$$

Where Δt is the diving time, Δx is the platform height and g=9.81 m/s^2 is the gravity.

Considering that a dive can have almost three or four figures inside, each of them can last almost:

$$t_{fig} = \frac{\Delta t}{5} = 0.1 \ [s] \tag{2}$$

That correspond to an acquisition rate of:

$$f = \frac{1}{t_{fig}} = 10 \ [Hz] \tag{3}$$

To be sure of avoiding aliasing, it is necessary to acquire at least at 20 Hz. To have a safety margin an acquisition rate of 25Hz has been used.

This is the acquisition rate from the video (usually recorded with a higher acquisition rate) to the frame.

The safety margin in the acquisition rate is also useful to have a redundancy of information to be able to face to problems in video analysis.

Considering that the total video time is approximately between 2 and 5 second, the total number of frame that should be analysed is between 80 to 200 frames. This number is acceptable because it can be processed without problems with a commercial PC, so the requirement of non-specialist equipment is satisfied.

After having done a sampling from the video, it is possible to analyse firstly all the frames with the mosaicking procedure and then frame by frame by the barycentre identification.

### 4.3. Mosaicking

Having extracted from the video the frames, the second step of the procedure is mosaicking.

Image mosaicking is a computational technique that exploit the presence of common features in a set of pictures [20] in images to a picture, called *panorama* [21].

Image mosaicking is a general [22] technique of image processing that has been applied to many fields, from medicine [23] to power plant inspection [24].

This technique is based on the identification of a set of features that are kept invariant during the most common transformations between pictures. The features of different pictures are matched and then the geometric transformation is identified. In this step to each picture a transfer function is associated [25].

At this point it is possible to wrap all the image in one image, called *panorama*. The *panorama* can be interpreted in this context as the common background to all the pictures.

In image mosaicking, it is possible to have different types of transformations [21]: testing many of them, it has been seen that the affine transformation is the best one for this application [26].

This step can solve two common problems related to the diving video: the first one is that the camera can have vibrations due to the low quality of the absence of an appropriate equipment. The second problem solved by mosaicking, is the fact that often, high quality video follows the diver, so the reference system changes (also a lot) from frame to frame.

> 1. Definition of the reference frame
> 2. Individuation of the main features of the reference frame
> 3. **For** all the frames **do**
> 4.     Identification of the main feature of the frame
> 5.     Feature matching between current frame and previous one
> 6.     Definition of an appropriate geometric transformation
> 7. **End For**
> 8, Definition of the size of the panorama
> 9. Empty panorama creation
> 10. **For** all the frames **do**
> 11.     Frame transformation in the global reference system
> 12.     Transformed frame added to the panorama
> 13. **End For**

Algorithm 1. Mosaicking procedure

Both these two problems can be easily solved by the mosaicking technique: a panorama is created and to each frame a transfer function is associated.

The result of the application of image mosaicking on a diving non-professional video in shown in Figure 2. In this case the technique has been used to eliminate vibrations in the movie.

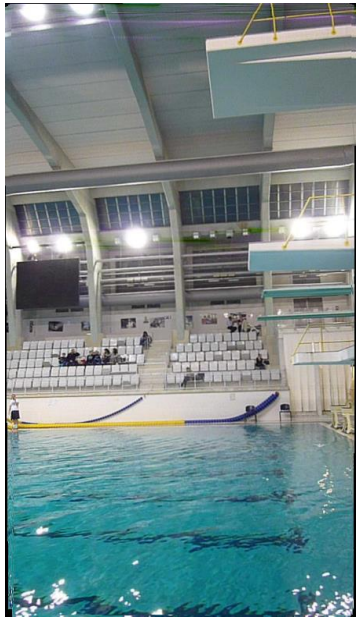

Figure 2. Example of image mosaicking for vibration elimination

In Figure 3 another example of application of the same procedure of image mosaicking has been applied on a video of a 10m platform dive of the Olympic Games of 2012. In this case the video has been recorded such that the athlete is almost al-ways at the centre of the video.

After having defined a background common to all the frames, it is possible to go on with the procedure of barycentre identification

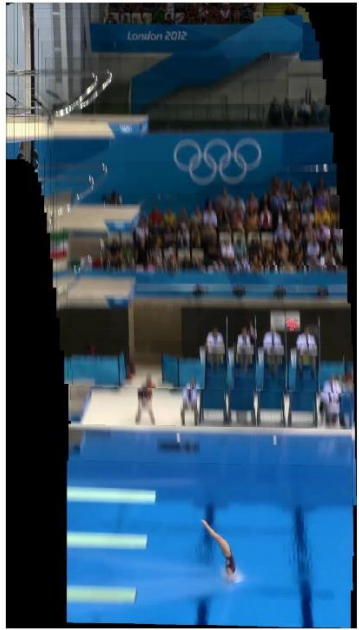

Figure 3. Example of image mosaicking for background reconstruction

## 4.4. Barycentre identification

Barycentre identification is a procedure that should be done frame by frame. In this paper, this procedure has been implemented using a proper colour filtering, as shown in Algorithm 2.

```
1. Filter parameters setting
2. Panorama filtering
3. For all the frames do
4.     Definition of the frame in the global reference system
5.     Frame colour filtering
6.     Definition of the difference between the filtered panorama and the filtered frame
7.     Object filtering
8.     Barycentre calculation
9. End For
```

Algorithm 2. Barycentre identification procedure

It is possible to make some comments on the barycentre identification procedure:

• Due to the flexibility requirement, the colour filter [27] has to be not much selective because the light changes from frame to frame, so the colour of the diver can change. Obviously, a non-selective filter let many parts of the background in the filtered image. To reduce the number of false positive recognitions, also the background obtained from the mosaicking is filtered: in this way, the parts of the background that passes through the filter are known and they can be eliminated from the filtered frame;

• The colour filtering has been done by double threshold function applied to each channel in an appropriate colour space [28]. After several trials, the most suitable colour space is the HSV colour space [29].

Figure 4 shows two examples of the barycentre identification: the area output of the colour filter is the one with the white contour. The calculated barycentre is represented by the red square.

Analysing Figure 4 it is possible to see that, even if the filter is not perfect due to the presence of the swimsuit and due to the darker illumination of the arms of the diver, the position of the identified barycentre is approximately true.

Figure 5 shows the barycentre vertical position in time. It has been reconstructed doing the barycentre identification procedure for all the frames of the diving.

The results of the barycentre identification have to be processed because are affected by the noise introduced by little mistakes in the colour filtering procedure. To improve it, a moving average filter has been applied: in this way, the noise is reduced or completely eliminated.

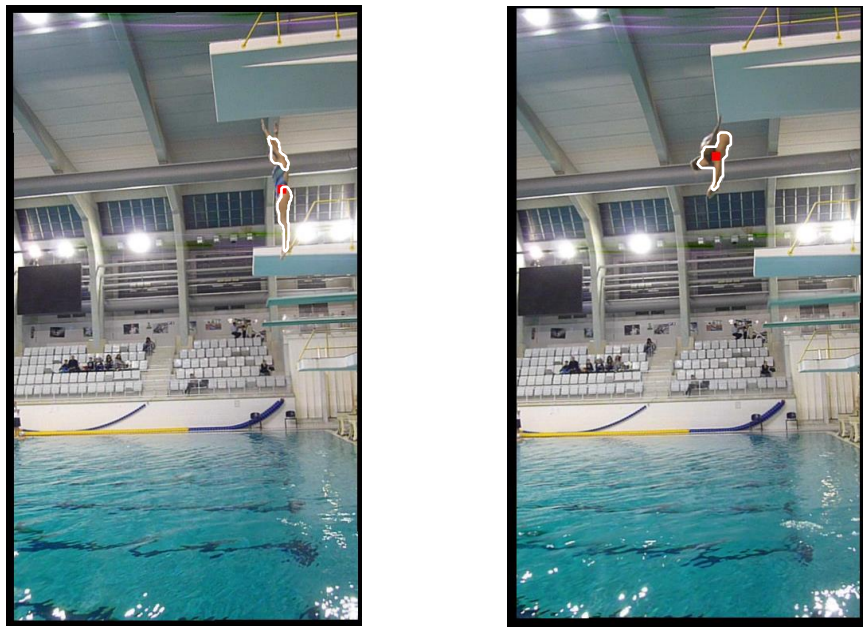

Figure 4. Example of barycentre identification

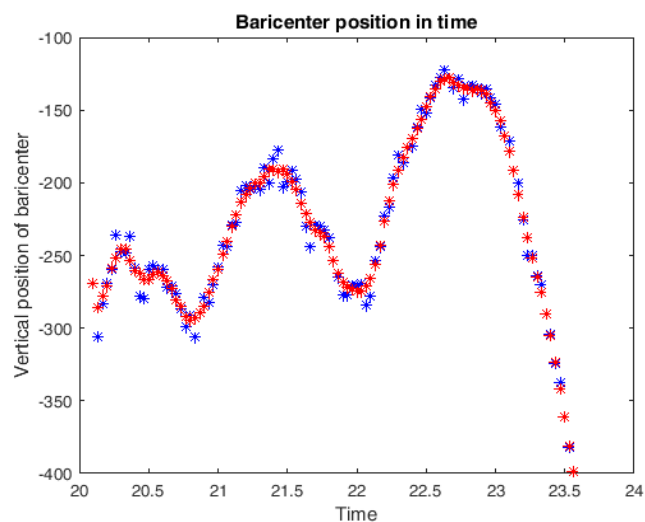

Figure 5. Barycentre vertical position in time. The blue dots are the positions directly taken from the analysis, while the red dots are the results of a filtering procedure

Figure 5 shows the comparison between the original position and the filtered ones: it is possible to notice that the filtered signal is able to reproduce correctly the large-scale movement of the athlete without introducing a delay. Moreover, the noise present in the original signal is correctly rejected.

## 5. CONCLUSIONS

In this paper a simple, economic and flexible procedure of video analysis of diving performance assessment has been presented. The technique is based on several steps that make the system as flexible as possible.

A possible improvement of the technique is the use of Deep Learning Neural Networks. Even if this last method is really powerful, it requires a huge number of in-put test cases: it is possible to apply the procedure described in this paper to prepare these inputs for the Network training.